\documentclass{amsart}

\usepackage[utf8]{inputenc}
\usepackage[english]{babel}
\usepackage{fullpage}
\usepackage{amsmath}
\usepackage{amssymb}
\usepackage{eufrak}
\usepackage{graphicx}
\graphicspath{{Figures/}}
\usepackage{subcaption}
\usepackage{hyperref,todonotes}
\usepackage{enumerate,color}
\usepackage{booktabs}
\usepackage{multirow}
\usepackage{multicol}
\usepackage[foot]{amsaddr}

\newcommand{\xslnote}[1]{}
\newcommand{\ignore}[1]{}

\newcommand{\TS}{\mathbb{TS}}
\newcommand{\IS}{\mathbb{IS}}
\newcommand{\OS}{\mathbb{OS}}

\newcommand{\DT}{\alpha}
\newcommand{\DI}{\beta}
\newcommand{\DO}{\gamma}
\newcommand{\NT}{\delta}
\newcommand{\NI}{\epsilon}

\newcommand{\T}{\mathit{T}}
\newcommand{\X}{\mathit{X}}
\newcommand{\Y}{\mathit{Y}}

\newcommand{\m}{{\it m}}
\newcommand{\n}{{\it n}}
\newcommand{\nodes}{{\it nodes}}
\newcommand{\cores}{{\it cores}}
\newcommand{\mb}{{\it mb}}
\newcommand{\nb}{{\it nb}}
\newcommand{\nth}{{\it nth}}
\newcommand{\nproc}{{\it nproc}}
\newcommand{\p}{{\it p}}
\newcommand{\q}{{\it q}}

\newcommand{\PDGEQRF}{{\it PDGEQRF}}
\newcommand{\MLA}{{{\it MLA}}}
\newcommand{\TLA}{{{\it TLA}}}

\DeclareMathOperator*{\argmin}{arg\,min}

\begin{document}

\title{Multitask and Transfer Learning for Autotuning Exascale Applications}

\author{Wissam M. Sid-Lakhdar}
\email{wissam@lbl.gov}
\address[1,3]{Lawrence Berkeley National Laboratory, Berkeley CA, 94720, USA}
\author{Mohsen Mahmoudi Aznaveh}
\email{aznaveh@tamu.edu}
\address[2]{Texas A\&M University, College Station TX, 77843 USA}
\author{Xiaoye S. Li}
\email{xsli@lbl.gov}
\author{James W. Demmel}
\email{demmel@berkeley.edu}
\address[4]{University of California, Berkeley CA, 94720 USA}

\maketitle

\begin{abstract}
Multitask learning and transfer learning have proven to be useful in the field of
 machine learning when additional knowledge is available to help a prediction task.
We aim at deriving methods following these paradigms for use in autotuning, where the goal is to find the optimal performance parameters of an application treated as a black-box function.
We show comparative results with state-of-the-art autotuning techniques.
For instance, we observe an average $1.5x$ improvement of the application runtime compared to the OpenTuner and HpBandSter autotuners.
We explain how our approaches can be more suitable than some state-of-the-art autotuners for the tuning of any application in general and of expensive exascale applications in particular.
\end{abstract}

\section{Introduction}
\label{sec:intro}

In 2016, the US Department of Energy (DOE) established the Exascale Computing Project (ECP, {\url{https://www.exascaleproject.org}}) to accelerate delivery of a capable exascale computing system.
In the ECP project, significant effort is being invested to develop highly scalable numerical libraries and high-fidelity modeling and simulation codes across the spectrum of science and engineering domains and disciplines.
Given a myriad of diverse computer architectures, achieving optimal performance and performance portability of all these codes has become an increasing challenge.
Each of these codes has a number of tuning parameters that may be difficult to choose to optimize performance (or any quantitative metric).
We are working on building an {\em adaptive execution} framework applicable across ECP applications.
Adaptive execution strategies are concerned with picking the right set of parameters to optimally solve a particular problem on a given architecture.
Our interrelated goals are to make it easy to do the following: 1) Support a variety of tuning metrics, including time, memory usage, accuracy, or hybrids like minimizing time given a memory constraint; 2) Compose large applications out of multiple code bases, including the ability to optimize the overall application, perhaps by changing data structures or the number of processors in between calls to different codes, or using different hardware resources; 3) Support archiving and reusing tuning data from multiple executions to allow tuning to improve over time; 4) Allow incorporation of new optimization techniques to improve the tuning process.

Since most of the exascale applications involve expensive ``function evaluations'', requiring either long runtime or many hardware resources (e.g., core count), the brute-force ``grid search'' approach to finding optimal parameters is infeasible.
Our approach is to build an automatic optimization routine to choose these parameters, using the application code as a ``black box'', and perform a small number of carefully chosen runs with different values of tuning parameters in order to find the best ones.
These parameters could be internal to one routine, or involve multiple routines that must be ``co-tuned'' because they share resources.
If a routine is independent of, or has a low impact on, the rest of the application in terms of the cost metric to be optimized, this routine can be tuned separately, providing the user can execute it in a stand alone black-box fashion.
Otherwise, the full exascale application needs to be run in order for the auto-tuning to proceed.

As subpart of the ECP project, we have built a common interface for autotuning with the following features:
1) Input data and parameters, or any statistics describing them, are exposed by the application to the optimizer.
These could include a pointer to prior information from a database, which we may also update with new data to improve the future adaptation layers;
2) Any resource constraints (e.g., core count or memory) are also exposed to the optimizer.
These could come from the application or a higher-level run-time system;
3) Optimization metrics and constraints (e.g., minimize runtime subject to a memory constraint) are also stated.
Note that this data could be the union of data provided by many components of a larger application.

The core idea, and the main contribution, of this paper is to broaden the scope of autotuning, contrasting with that of state-of-the-art approaches.
Specifically, instead of tuning a given application on a particular target problem instance, as has been done so far, we aim to tune that application for a myriad of problems it aims to solve.
The proposed realization of this idea is achieved through the use of {\it multitask} and {\it transfer} learning.
Experimental results show a $1.5x$ average improvement of the application runtime over state-of-the-art autotuning approaches, together with competitive (run-free) autotuning model predictions.
Moreover, we show that when the budget (number of allowed runs of the application) for autotuning is low, due to a high cost of the target application (e.g. exascale applications), our methods are more suitable than state-of-the-art methods that are not necessarily designed for this use case.

The remainder of this paper is organized as follows.
Section~\ref{sec:back} describes the related work in autotuning and black-box optimization for autotuning.
Section~\ref{sec:moat} presents the multi-output autotuning framework together with the notations used in this paper.
Section~\ref{sec:mtl} and~\ref{sec:tl} explain the new methodology that relies on multitask learning and transfer learning, respectively, for the purpose of autotuning.
Section~\ref{sec:exp} shows comparative results with the OpenTuner and HpBandSter state-of-the-art autotuners, together with the new results which are not possible to get through standard approaches alone.
Finally, section~\ref{sec:cncl} summarises this work and discusses its perspectives and future directions.

\section{Background}
\label{sec:back}

For a thorough survey of autotuning, multitask learning and transfer learning, we refer the reader to~\cite{8423171},~\cite{zy18} and~\cite{py10}, respectively.
Section~\ref{sss:bbo} describes black-box optimization as a mathematical framework for autotuning.
Black-box optimization techniques can be grouped in two categories.
{\it Model-free} or {\it non-Bayesian} optimization methods try to find an optimum through a deterministic and / or stochastic decision process.
Section~\ref{sss:nobayes} gives a brief overview of such methods, which are used in {\it OpenTuner}~\cite{opentuner} and {\it HbBandSter}~\cite{bohb}, two state-of-the-art autotuners that we compare our results with.
{\it Model-based} or {\it Bayesian} optimization methods build and optimize a surrogate model of the real objective function in order to find its optimum.
Section~\ref{sss:bayes} explains the principle of such methods, backbone of the machine learning methods of significance to our work, which are described later on in Section~\ref{sec:moat}, along with our proposed methodology.
Optimization methods that take into account application-specific knowledge are left aside to keep the focus on general purpose approaches instead.

\subsection{Black-box optimization for autotuning}
\label{sss:bbo}

The fundamental aspect of autotuning is optimization, i.e., finding a configuration of the parameters of an application that makes it solve a given input problem optimally.
The nature of autotuning makes this optimization problem lie within the family of black-box optimization problems, which are among the hardest to solve.
Indeed, an expensive run of the application is necessary in order to get the value of the objective function to optimize (e.g. runtime, energy) for a given combination of parameters.
Moreover, even in the presence of approximate (coarse) analytical models of the application (e.g. flop count, memory consumption), no precise enough formulation generally exists to predict more detailed phenomena (e.g. memory hierarchy communications, network contention) influencing the application behaviour.
Furthermore, the human cost of deriving such models (if even possible to derive) is too expensive to be practical.
Similarly, no analytical information on the gradient of the objective function is available, but only numerical approximations are, through sample evaluations of the objective function.

\subsection{Non-Bayesian optimization}
\label{sss:nobayes}

The simplest black-box optimization methods, that are usually tried first before resorting to more advanced methods, are known as: $(i)$ The deterministic {\it exhaustive search} (and its variant {\it grid search}), which tries all (or subset of all, respectively) possible combinations of all possible values of the parameters and selects the best performing one. Their drawback is that they quickly become intractable when the number of parameters increases, due to the curse of dimensionality~\cite{Bellman:1957}; $(ii)$ The stochastic {\it random search}, which selects randomly the value of each parameter in order to generate candidate solutions, then selects the best performing candidate.

Two main families of model-free optimization approaches exist.
The {\it global} approaches explore the whole search space and try to find a balance between exploration of new regions of the search space and exploitation of the data gathered to give more attention to the promising regions of space.
Examples of such methods are {\it Simulated Annealing} (SA)~\cite{sa}, {\it Genetic Algorithms} (GA)~\cite{ga} and {\it Particle Swarm Optimization} (PSO)~\cite{pso}.
In contrast, the {\it local} approaches try to improve upon previous solutions by exploring their neighboring region only until converging to a local minimum.
Examples of such approaches are {\it Nelder–Mead simplex}~\cite{NeldMead65} and {\it Orthogonal Search}~\cite{clp11}.

{\it OpenTuner}~\cite{opentuner} and {\it HbBandSter}~\cite{bohb} are two state-of-the-art general purpose autotuning frameworks.
They rely on meta-heuristics to solve a multi-armed bandit problem~\cite{kv87} where application runtime is the resource to be allocated (in our case).
OpenTuner allocates and distributes application runtime over a variety of optimization methods (mentioned in the previous paragraphs) in such a way as to adaptively select the best performing method, which is selected to solve the autotuning optimization problem.
HpBandSter's algorithm mixes a Bayesian Optimization method (see following section) with that of {\it Hyperband}~\cite{hyperband}.
Hyperband, as an early-stopping method, allocates runtime uniformly over randomly sampled configurations, keeping only the half best performing configurations at each iteration, and extending the corresponding runtime per remaining configuration, until a single (best) configuration remains.

Several other non-Bayesian black-box optimization packages for autotuning exist in the literature.
Particularly, SuRf~\cite{surf} uses random forests to model the performance of an application and find its optimum.
One of its main strengths is its ability to handle categorical parameters (choices) in an elegant way.

\subsection{Bayesian optimization}
\label{sss:bayes}

{\it Bayesian optimization}~\cite{sswaf16}, also known as {\it response surface methodology}, relies on a surrogate model of the real objective function to optimize.
A {\it prior} belief model representing the assumptions on the objective function is chosen, and a {\it posterior} is built from it so as to maximize the likelihood of some sample data of the objective function to be a realization of that model.
Instead of directly optimizing the true objective function, the model is optimized instead, as it is much cheaper to evaluate, and iteratively updated until convergence to an optimum.
The {\it Efficient Global Optimization} (EGO) algorithm~\cite{jsw98} is a classical Bayesian optimization algorithm.
In order to balance between exploration and exploitation (global and local search behaviors), EGO tries to optimize a certain acquisition function (e.g. {\it Expected Improvement} (EI)) instead of the model itself.
This metric considers both the value of the model at a given point in the search space together with the confidence in the model prediction at that location.
For instance, if the model predicts a good value of the objective function with a high confidence at a given location, while a worse value at another location but with a lower confidence, it might be worth exploring this second location as an optimum might be located nearby.
After finding a location in the search space that optimizes the acquisition function, this location is evaluated through the expensive black-box objective function and the corresponding value is used to update the surrogate model.
EGO iterates the process until the EI reaches a certain threshold, at which point the algorithm is considered to have converged to an optimum.

\section{Multi-Output AutoTuning} 
\label{sec:moat}

Throughout the paper, we refer to {\it input problem} as an input of the target application to be tuned, that is.
Also, we refer to {\it task} as the problem of tuning the parameters of the target application given a specific input problem.

Starting from the observation that tuning an application for a specific input problem makes that application efficient on solving that particular problem but without any guarantee on its behaviour on other unencountered problems, our aim in this paper is to tune an application for any problem it might encounter.
This goal being potentially unrealistic in general, given the infinity of input problems that might exist, our aim is to find, in a reasonable amount of time, good enough combinations of parameters of the application for any given input problem.

The guiding ideas behind our work are twofold.

\begin{itemize}
    \item 
Firstly, we tune the application on a finite set of well chosen input problems.
We believe that tuning the application on each problem independently from the others is less efficient than tuning all of them simultaneously, benefiting from the gathered knowledge on all input problems to speed up the whole tuning process
This is critical especially in an exascale setting, where every run of the application is extremely expensive.
We rely on the {\it multitask learning} framework to this end as detailed in Section~\ref{sec:mtl}.
    \item 
Secondly, we derive from the optimal parameters found on the initial set of problems, together with the model built on these problems, approximations to the optimal parameters of any other (non-tuned) input problem.
We rely on the {\it transfer learning} framework to do so as explained in Section~\ref{sec:tl}.
\end{itemize}

Let us define autotuning in the multi-output setting.
Throughout this paper, the {\it QR factorization} routine of {\it ScaLAPACK}~\cite{slug}, denoted as $\PDGEQRF$, is used as a guiding example of application to be tuned.
It takes as input a matrix $A$ and computes its QR factorization, producing an upper triangular matrix $R$ and a matrix $Q$.
Moreover, the notations used in this paper are summarized in Table~\ref{tab:smbl}.

\begin{table}[!h]
    \scalebox{1.}{
    \begin{tabular}{c|l|l|}
        \cline{2-3}
        & Symbol & Interpretation \\
        \cline{2-3}
        & \multicolumn{2}{|c|}{General notations}\\
        \cline{2-3}
        & $\TS$    & task space \\
        & $\IS$    & input space (parameter configurations) \\
        & $\OS$    & output space (e.g., runtime) \\
        \cline{2-3}
        & $\DT$    & dimension of $\TS$ \\
        & $\DI$    & dimension of $\IS$ \\
        & $\DO$    & dimension of $\OS$ \\
        & $\NT$    & number of tasks \\
        & $\NI$    & number of samples per task \\
        \cline{2-3}
        & $\T \in \TS^{\NT}$ & matrix of tasks selected from sampling \\
        & $\X \in \IS^{\NT \times \NI} $ & matrix of samples (parameters)\\
        & $\Y \in \OS^{\NT \times \NI} $ & vector of results (runtime) \\
        \cline{2-3}
        & \multicolumn{2}{|c|}{ScaLAPACK $\PDGEQRF$ related notations}\\
        \hline
        \multicolumn{1}{|c|}{\multirow{4}{*}{\rotatebox[origin=c]{90}{Task}}}
                               & $\m$     & number of matrix rows \\
        \multicolumn{1}{|c|}{} & $\n$     & number of matrix columns \\
        \multicolumn{1}{|c|}{} & $\nodes$ & number of compute nodes \\
        \multicolumn{1}{|c|}{} & $\cores$ & number of cores per compute node \\
        \hline
        \multicolumn{1}{|c|}{\multirow{6}{*}{\rotatebox[origin=c]{90}{Input}}}
                               & $\mb$    & row block size \\
        \multicolumn{1}{|c|}{} & $\nb$    & column block size \\
        \multicolumn{1}{|c|}{} & $\nproc$ & number of MPI processes \\
        \multicolumn{1}{|c|}{} & $\nth$   & number of threads (used in BLAS) \\
        \multicolumn{1}{|c|}{} & $\p$     & number of row processes \\
        \multicolumn{1}{|c|}{} & $\q$     & number of column processes \\
        \hline
    \end{tabular}
    }
    \caption{Symbol table}
    \label{tab:smbl}
\end{table}

Let us define $\TS$, the {\it Task Space}, as the space of all the input problems that the application may encounter.
Let us make the simple (non restrictive) assumption that $\TS$ may be characterised as a finite dimensional space of dimension $\DT$.
This means that it is possible to identify any input problem with a finite number of features.
Relating to $\PDGEQRF$, $\TS$ is a space of dimension $\DT = 4$.
The corresponding features are: $\m$ and $\n$, respectively the number of rows and columns of a matrix to be factorized; $\nodes$ and $\cores$, respectively the number of compute nodes together with the number of processor cores per compute node, characterizing the architecture of the machine on which the application runs.
Several application problems are amenable to such a formalism, potentially after some approximations are made.
If this finite dimension task space assumption does not hold, the multitask learning methods in Section~\ref{sec:mtl} remain valid, but some of the transfer learning methods in Section~\ref{sec:tl} would require additional attention that we leave as a future work.
An example of application where this assumption does not hold is the case of {\it SuperLU}~\cite{lidemmel03,Li05}, where the task space represents the space of all possible sparse matrices, which cannot be characterized by a finite set of parameters. 
The underlying assumption that we make is that the objective function to optimize is somehow continuous in the task space, as we expect it to be similar for similar tasks (for the same parameters values).

Let us define $\IS$, the {\it Input Space}, or space of the parameters to be optimized, as $\mathbb{R}^{\DI}$, with $\DI$ the number of parameters.
In practice, inputs can be either real, integer or categorical (e.g. a list of $n$ algorithms (which can map to an interval $[1,n]$)).
In our work, the last two cases are translated internally to the real case.
Every point in $\IS$ can be referred to as a {\it parameter configuration}.
Moreover, most applications have constraints on their parameters as not all possible combinations of parameters (point in the input space) are valid.
Relating to $\PDGEQRF$, $\IS$ is a space of dimension $\DI = 6$.
The corresponding parameters are: $\mb$ and $\nb$, the row and column block sizes, respectively; $\nproc$ and $\nth$, the number of MPI processes and BLAS threads, respectively, and $\p$ and $\q$, the number of row processes and column processes, respectively.
Moreover, two constraints on these parameters exist: ($i$) $\nproc = \p \times \q$; ($ii$) $\nproc \times \nth = \nodes \times \cores$.
These constraints reduce the effective dimension of the input space to $\DI = 4$, as $\nth$ and $\q$ can be deduced from the other parameters.

Let us define $\OS$, the {\it Output Space}, as the space of the results of the evaluation of the optimization objective function, for a given task and on a given parameter configuration.
It is a single dimensional space (e.g., computation time, memory consumption, energy \dots), the case of multi-dimensional spaces (corresponding to multi-objective optimization) being left for future work.

We denote by $y(t,x) \in \OS$ and $f(t,x) \in \OS$ the values of the objective function measure $y$ and model prediction $f$, respectively, for a task $t \in \TS$ and for a parameter configurations $x \in \IS$.
As is the case in Bayesian optimization, the model $f$ is optimized instead of the measured value $y$, while the optimum found is hoped to be that of $y$.
In this setting, given a task $t \in \TS$, the autotuning goal is to find:
\begin{equation}
    \argmin_{x \in \IS} f(t,x)
\end{equation}
The autotuning stopping criteria is classically defined as one or a combination of the following: $(i)$ a maximum number of evaluations of the objective function (runs of the application); $(ii)$ a maximum wall-clock time; $(iii)$ threshold on a relevant measure of the quality of the solution provided by the application (e.g., numerical accuracy, energy consumption).
In order to be able to fairly compare different autotuning algorithms, we chose to fix a maximum number of runs of the application as the stopping criteria.
The total runtime spent in the application is also measured and reported.

\section{Multitask learning}
\label{sec:mtl}

In the field of machine learning, multitask learning consists of learning several tasks simultaneously while sharing common knowledge between them in order to improve the prediction accuracy of each task and / or speed up the training process.
In this section, we propose the {\it Multitask Learning Autotuning} method, $\MLA$. 
The methodology followed in $\MLA$ resembles that of single-task Bayesian optimization methods, as it adapts the EGO algorithm to the multi-output setting.
It is composed of three main phases: $(i)$ sampling phase (Section~\ref{ss2:smpl}); $(ii)$ modeling phase (Section~\ref{ss2:mdl}); and $(iii)$ optimization phase (Section~\ref{ss2:opt}).
The second and third phases are repeated until convergence or stopping of the tuning process.

\subsection{Sampling phase}
\label{ss2:smpl}

While a single sampling step is needed in a single-task Bayesian optimization scheme, two sampling steps are needed in $\MLA$.

The objective of the first sampling step is to select a set $\T$ of $\NT$ tasks $\T = [t_{1}; t_{2}; \dots; t_{\NT}] \in \TS^{\NT}$.
The goal when selecting this set is to get a representative sample of the variety of problems that the application may encounter, rather than focusing on a specific type of problem.
Given the freedom in the selection of the tasks together with the existence of a space of tasks $\TS$, we chose a {\it space filling sampling} in $\TS$ to select the $\NT$ tasks.
Such samplings are widely used in the field of {\it Design Of Experiments} (DOE)~\cite{doe}.
Particularly, we chose a {\it Latin Hypercube Sampling} (LHS)~\cite{lhs} in our method.
Such samplings try to cover the whole search space uniformly, such that no region of the space is over-sampled while another one is left under-sampled.
Several off-the-shelf software packages exist that implement several different types of samplings (including LHS).

The objective of the second sampling step is to select an initial sampling $\X$ of parameter configurations for every task $\X = [X_{1}; X_{2}; \dots; X_{\NT}] \in \IS^{\NT \times \NI}$. 
For task $t_{i}$, its initial sampling $X_{i}$ consists of $\NI$ parameter configurations $X_{i} = [x_{i,j}]_{j \in [1,\NI]} \in \IS^{\NI}$.
Two cases arise in the multitask framework: the {\it isotropic} and {\it heterotropic} cases.
The isotropic case arises when all the tasks share the same sampling in the input space, while in the heterotropic case, different tasks do not necessarily share the same samples.
In the isotropic case, the advantage of a multi-output regression is the sharing of information for the optimization of the hyper-parameters of the model governing the tasks.
In the heterotropic case, however, more knowledge can be shared as insights on the true cost of a task on an unknown configuration can be learned from a similar task with a sample at that location.
Moreover, in real-life applications, given the existence of constraints on the parameters, not all parameter configurations are feasible for all tasks simultaneously; a configuration may be valid for a subset of tasks, but violate the constraints on another subset.
Thus, we chose to generate the initial sampling $X$ in a heterotropic way by generating the $X_{i}$ as independent LHS.

Every sample $x_{i,j}$ is evaluated through a run of the application, whose result is represented as $y_{i,j} = y(t_{i}, x_{i,j}) \in \OS$.
The set $\Y$ represent the results of all these evaluations, $\Y = [Y_{1}; Y_{2}; \dots; Y_{\NT}] \in \OS^{\NT \times \NI}$, where every $Y_{i}$ represents the results corresponding to task $t_{i}$, $Y_{i} = [y_{i,j}]_{j \in [1,\NI]} \in \OS^{\NI}$.

Given the application constraints, a generic sampling technique might fail, both for the selection of the task samples and for the selection of their corresponding input samples.
Such a case arise frequently when the total number of combinations of parameter values is of the order of thousands of billions (simply because of the curse of dimensionality) while the number of valid parameter configurations that respect the constraints is only of the order of hundreds of thousands.
An example is the tuning of matrix multiplication on GPU in MAGMA~\cite{ahkld15,magma}.

In such a case, either specific knowledge of the application should be used to design a space filling sampling, or, a simple remedy is to generate several unconstrained samplings and randomly select candidates from the union of sets of valid configurations drawn from every sampling.

\subsection{Modeling phase}
\label{ss2:mdl}

Once $\T$ and $\X$ are selected and $\Y$ evaluated, the core of $\MLA$ is its modeling phase, which consists in training a model of the black-box objective function relative to tasks $\T$.
However, instead of building a separate model for every task, as is customarily the case in a regular single-task Bayesian optimization scheme, the challenge in $\MLA$ is to derive a single model that incorporates them all, sharing the knowledge between them to be able to better predict them all.
While {\it Gaussian Processes} (GPs) are often used in the modeling for single task tuning (e.g. ~\cite{spearmint}), we propose to rely in $\MLA$ on the generalization of GPs to the multi-output setting, known as the {\it Multi-Output Gaussian Process} framework~\cite{bkw08}.
Particularly, we chose to use the {\it Linear Coregionalization Model} (LCM)~\cite{nla.cat-vn1116468,CIS-7228}, the choice of which is driven by its generality, flexibility and modeling power, albeit its modeling cost, compared to the plethora of models in the literature that are derived from it as special and constrained cases~\cite{bkw08}.

A similar but orthogonal line of work that also relies on an agglomerate model of tasks is presented in \cite{fsk07}.
Indeed, in the context of multi-fidelity optimization, where one seeks the optimum of an expensive task while a hierarchy of highly correlated and cheaper tasks exist, the authors show that the use of {\it co-Kriging} (which can be formulated as a special case of LCM) leads to a more accurate model, and subsequently, to a faster convergence to a global optimum of the expensive task.
However, instead of a (vertical) hierarchy of correlated tasks of increasing cost, our problem corresponds to a (horizontal) relationship between correlated tasks of the same (or similar) cost. 

After describing the concept of GPs in section~\ref{ss2:gp}, we describe the LCM in section~\ref{ss2:lcm}.

\subsubsection{Gaussian Processes}
\label{ss2:gp}


We provide here a brief explanation of Gaussian processes.
We invite the reader to consult~\cite{rw05} for a detailed description. 
A Gaussian process is the generalization of a multivariate normal distribution to an infinite number of random variables.
It is a stochastic process where every finite subset of variables follows a multivariate normal distribution.
While other regression methods set a prior on the function to be predicted and try to learn the parameters of such a function, GPs set a prior on some characteristics of the functions (e.g. smoothness) and try to learn the functions themselves, allowing for the expressiveness of a much richer variety of functions.
A GP is completely specified by its mean function $\mu(x)$ and by its covariance function $k(x,x')$.
A function $f(x)$ following such a GP is written as:

\begin{equation}
    f(x) \sim GP(\mu(x), k(x,x'))
\end{equation}

where:

\begin{equation}
    \mu(x) = \mathbb{E}[f(x)]
\end{equation}

\begin{equation}
    k(x,x') = \mathbb{E}[(f(x) - \mu(x))(f(x') - \mu(x'))]
\end{equation}

In most practical scenarios, $\mu$ is taken to be the null function and all the modeling is done through $k$, known as the kernel function.
While a variety of kernel functions exist in the literature, the choice of which is dependant on the data to be modeled, a very popular generic choice is the exponential quadratic kernel given by:
\begin{equation}
    k_q(x,x') =  \sigma_{q}^{2} \exp \Bigg(-\sum_{i=1}^{\DI}\frac{(x_{i} - x'_{i})^{2}}{l_{i}^{q}} \Bigg)
    \label{eq:kq}
\end{equation}
where $\sigma_{q}^{2}$ (variance) and $l_{i}^{q}$ (length scales) are hyper-parameters of the kernel governing its behavior.
These are learned by optimizing the log-likelihood of the samples $X$ with values $y$ on the GP, the log-likelihood being described as:

\begin{equation}
    \begin{split}
    log(p(y|X)) =& -\frac{1}{2} (y - \mu(X))^{T}(K + \sigma^{2}I)^{-1}(y - \mu(X))\\
                 & - \frac{1}{2} log|K + \sigma^{2}I| - \frac{n}{2} log(2\pi)
    \end{split}
\end{equation}

where $\sigma^{2}I$ is a regularization term, and $K$ is the covariance matrix whose elements are generated from the kernel $k$.

\subsubsection{Linear Coregionalization Model}
\label{ss2:lcm}

The key to LCM is the construction of an approximation of the covariance between the different outputs of the model (model of every $t \in T$).


In this method, the relations between outputs are expressed as linear combinations of independent {\it latent random functions}
\begin{equation}
    f(t_{i},x) = \sum_{q=1}^{Q} a_{i,q} u_{q}(x)
    \label{eq:fi}
\end{equation}
where $a_{i,q}$ ($i \in [1,\NT]$) are hyper-parameters to be learned, and $u_{q}$ are the latent functions, whose hyper-parameters need to be learned, as well.

The independence hypothesis is important as it allows us to compute the covariance between the outputs only through the auto-covariance of the latent functions themselves, as it implies:
\begin{equation}
    cov(u_{i}, u_{j}) = 
      \begin{cases}
        cov(u_{i}, u_{i}) &, if~i = j\\
        0                 &, if~i \neq j
      \end{cases}
  \label{eq:covind}
\end{equation}

The covariance of every latent function $u_{q}$ is assumed to be generated from a kernel function:
\begin{equation}
    cov(u_{q}(x), u_{q}(x')) = k_q(x,x') 
    \label{eq:covkq}
\end{equation}
This kernel can take $x$ and $x'$ to be vectors as input, in which case, $k_q$ is a scalar.
However, this kernel can also consider them to be vectors of vectors (i.e.: matrices).
In such a case, $k_q(x,x')$ is a matrix, instead of a scalar, where the $(i,j)$ entry corresponds to the evaluation of the kernel $k_q$ on the $i^{th}$ vector of $x$ and $j^{th}$ vector of $x'$.

The covariance structure between two outputs can now be expressed as:
\begin{equation}
    cov(f(t_{i},x), f(t_{i'},x')) = \sum_{q=1}^{Q}\sum_{q'=1}^{Q} a_{i,q}a_{i',q'} cov(u_{q}(x), u_{q'}(x'))
\end{equation}
which, thanks to Equation~(\ref{eq:covind}) simplifies to:
\begin{equation}
    cov(f(t_{i},x), f(t_{i'},x')) = \sum_{q=1}^{Q} a_{i,q}a_{i',q} cov(u_{q}(x), u_{q}(x'))
\end{equation}

The covariance matrix between all the tasks on inputs $x$ and $x'$ can now be expressed through the kernel $K(x,x')$
\begin{equation}
    K(x, x') = \sum_{q=1}^{Q} B_{q} \otimes k_{q}(x,x') + D \otimes I
\end{equation}
where $\otimes$ is a Kronecker product, $k_{q}$ is the covariance function of the $q^{th}$ latent function, $I$ is the identity matrix of size $\NI \times \NI$, $D$ is a diagonal matrix, whose elements are the variance of the noise in the measurement of the samples, and $B_{q}$ is a $\NT \times \NT$ matrix of parameters of the model such that:
\begin{equation}
    B_{q}[i,i'] = a_{i,q}a_{i',q} = W_{q} W_{q}^{T}
\end{equation}
with $W_{q}$ a vector of parameters.
The $D \otimes I$ term acts as a regularization term that prevents overfitting and helps make the covariance matrix non-singular.


The parameters that need to be learned are the elements of the $W_{q}$ vectors as well as the hyper-parameters of the kernels $k_{q}$.


In order to find the best hyper-parameters of the model, the log-likelihood of the model on the data needs to be optimized.
The solution of this non-convex optimization problem is a subject of active research.
Some software libraries rely on model-free black-box optimization techniques to solve it while others rely on gradient-based optimization techniques with multi-start (to circumvent the non-convexity), as is the case in the {\it GPy}~\cite{gpy} on which we rely in our work.

\ignore{
\subsubsection{Deep Gaussian Processes}

In practice, discontinuities can occur. A more general assumption we could make instead would be to consider the objective function to be piece-wise continuous.  
In this study however, we restrict ourselves to the continuous case.

The only occurrence of the use of deep GPs for autotuning in the literature that we are aware of is the work of \dots in~\cite{XXX}.

The pros of deep GPs make it a great tool for tuning exascale applications, while its usual computational drawbacks are leveraged in the exascale setting where few samples only can be collected, making the deep GP model tractable (albeit some approximations).
}

\subsection{Optimization phase}
\label{ss2:opt}

Once the model of $\MLA$ is either built (at the first iteration of $\MLA$) or updated (at subsequent iterations), the optimization phase consists of finding the optima of an acquisition function over the model (such as EI), relatively to every task, through a model-free black-box optimization algorithm (such as {\it Particle Swarm Optimization} (PSO)).
The resulting solution found for every task is then evaluated through an expensive run of the application to be tuned and then re-injected into the model.

The model can be queried for a prediction relative to a task $t_{i}$ and for an (unexplored) input $x^{*}$, using not only the data of the task itself but also the data of the other tasks, through the formulation:
\begin{equation}
    f(t_{i},x^{*}) = K(x^{*}, X)^{T} K(X,X)^{-1} Y
    \label{eq:pred}
\end{equation}

A fundamental property of GP-based models (such as LCM) is the ability to estimate the confidence in the predictions alongside the predictions themselves.
The confidence can be expressed as:
\begin{equation}
    var(t_{i},x^{*}) = K(x^{*}, x^{*}) - K(x^{*}, X)^{T} K(X,X)^{-1} K(x^{*}, X)
    \label{eq:var}
\end{equation}

The formulation in Equations~(\ref{eq:pred}) and~(\ref{eq:var}) holds for both the GP and LCM models.

There are several benefits to using a multi-output framework during the optimization phase.

\begin{enumerate}
    \item
The independence of the different optimizations (of the different tasks) allows for a high degree of parallelism, that can easily be exploited.
This comes in addition to the parallelism available within the EGO algorithm itself.
    \item
The optimum of every task should be close to the optimum of related (close) tasks.
Thus, the exploration around the optimum of one task will benefit the neighboring tasks as new data in the vicinity of their own optimum is gathered.
    \item
The effect of taking into account other tasks while predicting a given task increases the confidence in the prediction (decreases variance), which leads the $\MLA$ to converge faster to the global optimum.
\end{enumerate}

\section{Transfer learning}
\label{sec:tl}
In the field of machine learning, transfer learning consists of using the knowledge of one (or several) task(s) to improve the learning accuracy and / or speed of another task.
Given the knowledge gathered and model built on previously autotuned representative tasks, we would like to be able to predict an optimal, or at least good enough, parameter configuration on a new unknown task.
To this end, we propose two novel methods, the second one building on the first one.
By combining in a new way some basic building blocks such as sampling techniques and GPs, they are able to target the needs of the multi-output autotuning framework that we develop in this paper.

The first method, that we denote as $\TLA1$ (short for {\it Transfer Learning for Autotuning 1}) and describe in Section~\ref{ss2:mdlopt}, applies if an approximation to the optimum parameter configuration of a new task is considered enough, or, if a specific tuning for that task cannot be afforded.

The second method, that we denote as $\TLA2$ and describe in Section~\ref{ss2:newtsk}, applies if a high(er) quality of the optimal parameter configuration of a new task is desired and a quick additional tuning step can be afforded.

Let us note that both $\TLA1$ and $\TLA2$ benefit from each other.
While $\TLA2$ relies on the approximation provided by $\TLA1$ to speed up its tuning, $\TLA1$ also benefits from the fact that $\TLA2$ updates the $\MLA$ model described in Section~\ref{ss2:mdl} every time it is applied on a new task.

\subsection{$\TLA1$: Modeling the optima}
\label{ss2:mdlopt}

The goal of the $\TLA1$ method that we propose is to create a model that can predict the optimal configuration corresponding to an unexplored task without having to tune it.

Given the tuning of the set of tasks $T$, let us define the set of corresponding optima $OPT$ as
\begin{equation}
    OPT_{i} = \argmin_{x \in \IS} f(t_{i},x), \forall i \in [1,\NT]
\end{equation}

An optimum parameter configuration is composed of as many parameters as the dimension of the input space $\IS$, i.e. $\DI$.
Consequently, the solution that we propose is to create $\DI$~separate and independent Gaussian Processes $G_{i \in [1,\DI]}$ to model every component of the optimal solutions separately.
Such a Gaussian Process model is described in Section~\ref{ss2:gp}.
Relating to $\PDGEQRF$, $G_{1}$ predicts $\mb$ and $G_{2}$ predicts $\nb$, for example.
The set $OPT$ represents the input data of every one of these Gaussian Processes.
It is important to notice that, while the model in Section~\ref{ss2:mdl} has its space of inputs $\IS$ and space of outputs $\OS$, every one of the Gaussian Processes created here has its space of inputs $\TS$ and its space of outputs in one of the dimensions of $\IS$.

For any unexplored task $t^{*}$, a prediction of its optimal parameter configuration is then given by
\begin{equation}
    OPT(t^{*}) = [G_{1}(t^{*}), G_{2}(t^{*}), \dots, G_{\DI}(t^{*})] \in \IS
\end{equation}

Moreover, the confidence in the prediction of the $G_{i}$ models can serve as an indicator to the user of the autotuning on whether an additional tuning step (as described in the next section) is needed or not.

An alternative solution could have been to define a single multi-output Gaussian Process to model all the components simultaneously.
However, no a priori hypothesis can be made in the general case on the correlations between the different components characterising the optima.

\subsection{$\TLA2$: Adding a new task}
\label{ss2:newtsk}

The goal of the $\TLA2$ method that we propose is to speed up the tuning of an unknown task by leveraging the knowledge of previously tuned tasks.
It is composed of two steps.

\xslnote{Can the description of ``first step'' be shortened? since it is related to TLA1 ? ==> wissam: the method in this section uses the result of TLA1 but is completely new and need to be described}

\subsubsection{Step 1: Centered sampling}

The first step consists in predicting the optimum of the new task (through $\TLA1$) in order to focus the attention of the tuning to its surrounding area.
Instead of sampling the whole search space of the new task with a general space filling sampling, a denser sampling is achieved in the area at the vicinity of the predicted optimum, while further away regions of the search space are less sampled.
Specifically, given a budget of a certain number of initial samples, a normal distribution is used to map these samples on the search space.
The center of the distribution is the predicted optimum.
The standard deviation of the distribution is chosen as the diameter of the search space.
Generated samples that are outside of the search space or violating constraints on the parameters are simply discarded and replaced by valid newly generated samples.

Let us note that, in the literature, the proposed sampling method is usually avoided.
If used in conjunction with a regular Gaussian Process to model the objective function of the new task without any additional knowledge from other tasks, the EGO algorithm applied on this Gaussian Process will tend to favor the exploration of the under-sampled areas, characterized by a high(er) prediction uncertainty.
However, this would be unnecessary in our case as the model used in the second step helps circumvent this situation as the knowledge on the previously autotuned tasks informs the EGO algorithm of the low probability of finding an optimum in the low sampled regions of the space.
Indeed, given the hypothesis of smoothness and continuity of the objective function between neighboring tasks, there should be no need to explore regions of the input space outside of the region neighboring the approximation to the optima.

A future improvement of the sampling is to chose a different variance of the distribution per dimension of the space.
The lengthscale hyper-parameter ($l_{i}^{q}$ terms in Equation~(\ref{eq:kq})) of the latent functions of the model of $\MLA$ relative to every dimension of the search space can be used to figure out a relevant variance of the sampling distribution for that dimension.
This improvement is simply inspired by {\it Automatic Relevance Determination} (ARD) methods~\cite{ard} where a relatively high value of the lengthscale hyper-parameter $l_{i}^{q}$ relative to one dimension is an indicator of a small variance in that dimension, and vice-versa.


\subsubsection{Step 2: Extended model}

The second step consists of extending the model of Section~\ref{ss2:mdl} to account for the new task to be tuned.
Specifically, given the formulation of the model of a given task described in Equation~(\ref{eq:fi}), the $B_{q}$ matrices need to be extended by one more row and one more column.
Given the formulation in Equation~(\ref{eq:kq}), this boils down to extending the vectors $W_{q}$ by a single additional scalar (hyper-parameter), for all $q \in [1,Q]$.
This modified model has two computational advantages.

Firstly, instead of relearning all of the hyper-parameters of the model, only the $Q$ additional hyper-parameters need to be learned, the other ones remaining unchanged.
Indeed, the elements of the $W_{q}$ vectors describe the relation between the latent functions $u_{q}$ and the task models $f(t_{i})$.
Adding a new task should not affect this relation (too much) on the previously tuned tasks, and, even if it does, these tasks being already tuned, no learning is needed on them, and the transfer of the learning is meant to be only from them to the new task.

Secondly, the costliest part of the training and building of the model of both $\MLA$ and $\TLA2$ is the inversion of the covariance matrix, as is the case with any Gaussian Process-based method.
While a full factorization of the whole covariance matrix is needed in $\MLA$, efficient {\it update / downdate} techniques~\cite{Davis09} can be applied in $\TLA2$.
These techniques consist of reusing the factors of the previously factorized part of the covariance matrix and only inverting the newly added rows and columns of this matrix.
In the case of $\TLA2$, these new rows and columns correspond to the Kronecker product of the new rows and columns of the $B_{q}$ matrices with the covariance matrices generated by the $k_{q}$ kernels on the samples of the new task.
The use of update / downdate techniques reduces the model building and training dramatically as they reduce the complexity of such operations from $O((\NT\NI)^{3})$ to $O(\NT(\NI)^{3})$.



\subsection{$\TLA2$ with restarting}

The guiding idea behind $\TLA2$ is that an expensive model is to be built once only, in $\MLA$ on an initial set of (training) tasks, while every additional unencountered task can be tuned at a much reduced cost, by taking advantage of it and by only updating it rather than by recomputing it from scratch.
However, this gain in model training cost comes at the expense of a potential loss in model accuracy.

While $\TLA2$ only adds and optimize new hyper-parameters in the $W_{q}$ vectors, it does not modify the existing hyper-parameters in these $W_{q}$ vectors, nor does it change the hyper-parameters of the kernels related to the latent functions $u_{q}$, nor does it increase the number of latent functions $Q$.
In this regard, one can view the optimization strategy in $\TLA2$ as being analogous to stochastic gradient decent while that of $\MLA$ would be analogous to gradient decent.

However, while it is a reasonable approximation to keep the old hyper-parameters of the model fixed when adding few new tasks (as in $\TLA2$), it might become advantageous to update these hyper-parameters if several new tasks are added, or even necessary to rebuild a new model from scratch, with a larger number of latent functions $Q$, if too many new tasks are added (as in $\MLA$).

The choice of whether to continue relying on $\TLA2$ when a yet another new task is to be tuned rather than {\it restarting } $\TLA2$, using $\MLA$ instead, is highly dependant on: the application to be tuned; the tuning quality sought for; and the tuning cost a user is willing to pay.
Specifically, two extreme regimes can be identified in autotuning.
In the first extreme regime, such as tuning exascale application, every run on the objective function is very costly.  Hence, the amount of data available to the tuner is small and every data is valuable.
In such a context, the cost of optimizing the $\MLA$ model is marginal.
It is thus better for a user to restart $\TLA2$ every time a new task is to be tuned.
In the second regime, every run of the application is cheap.  Hence, the amount if data available is considerable.  The cubic complexity of the inversion of the covariance matrix in the $\MLA$ model would be a bottleneck to the tuning.  It is then better for a user to rely on $\TLA2$ without any restart.
For a given application, the need to restart $\TLA2$ is then depending on the cost of $\MLA$ relative to the cost of a run of the application.








\section{Experimental results}
\label{sec:exp}

This section presents the experimental results of the methods described in this paper together with their comparison with state-of-the-art autotuning methods.
The autotuning of the {\it QR factorization}~\cite{demmel2012communication} routine $\PDGEQRF$ in the {\it ScaLAPACK} library~\cite{slug} is explored as a guiding example.

We use the ``Edison'' machine at NERSC.%
\footnote{\url{http://www.nersc.gov/users/computational-systems/edison/}}
Each node of Edison contains dual-socket 12-core Intel Ivy Bridge processors.
The code was compiled with the Intel Fortran compiler version 18.0.1 and linked with cray-libsci 18.03.1 for BLAS and ScaLAPACK operations.
 
The implementation of the algorithms proposed in this paper is done using the Python language.
Also, the code relies on the {\it GPy} library~\cite{gpy} from University of Sheffield, which implements a general framework for Gaussian Processes.

Finally, to cope with runtime instabilities and noise in the measurements, all the runs of the application were performed $3$ times, and the minimal runtime was selected.

\subsection{Analytical model}

Although the methods in this paper do not rely on any analytical model but only rely on machine learning models, and, given the availability of such analytical model in the case of the QR factorization, we show here such models, although bringing insights about the behaviour of the application sought to be tuned, are not precise enough to be reliable models for autotuning, which highlights the importance of the use of black-box machine learning models instead.

Detailed analytical models of QR factorizations runtime have been derived in the literature, some of which target distributed-memory algorithms.
In this section, we show that such models are not accurate enough to predict
performance.

In our study, we used three models.
All of them estimate the runtime as a linear combination of factors and model parameters as follows:
\begin{equation}
    Runtime = C_f \times t_f + C_m \times t_m + C_v \times t_v
\end{equation}
The factors represent the number of floating point operations $C_f$, the number of messages $C_m$ and the volume of data transferred $C_v$.
The model parameters relative to each factor are the time per floating point operation $t_f$, the time per message $t_m$ and the time per data item communicated $t_v$.
They are hardware dependent and need to be estimated through a linear regression procedure.

The models differ in the varying levels of complexity and accuracy of the estimates of the factors they rely on.
The first (basic) model is from \cite{slug}.  
It relies only on the input matrix size and number of processes
 to compute the factors.
The second (medium) model is formulated in \cite{demmel2012communication}.
It separates the effect of the number of divisions from the rest of the floating point operations.
Its estimate of the factors is given by:
\begin{equation}
    \label{eq:paef2}
    \begin{aligned}
        \#flops     &= \frac{2n^2(3m-n)}{3nproc}+\frac{bn^2}{2q}+\frac{3bn(2m-n)}{2p}+\frac{b^2n}{3p}\\
        \#divisions &= \frac{mn-n^2/2}{p} \\
        \#messages  &= 3n\log{p}+\frac{2n}{b}\log{q}\\
        \#words     &= (\frac{n^2}{q}+bn)\log{p}+(\frac{mn-n^2/2}{p}+\frac{bn}{2})\log{q}\\
    \end{aligned}
\end{equation}
where $b$ represents the block size (square blocks with $b = mb = nb$).
The third (advanced) model given by~\cite{Laura} consists of using a MATLAB simulation that takes into account more detailed aspects of the factorization algorithm (particularly the [boundary conditions / edge cases]) in order to compute an accurate count of the factors in Equation~(\ref{eq:paef2}).

In order to estimate the model parameters, we ran the QR factorization on a set of square matrices of size $1000$, $2000$ and $4000$, and chose the values of the $mb, nb, nproc, nth, p, q$ parameters following a semi-exhaustive (grid) search.
Given that the analytical models that we found in the literature consider only square blocks ($mb=nb$), only experiments involving square blocks (between $200$ and $600$ instances) are used for the fitting, the remainder of the data (between $6000$ and $22000$ instances) is kept for model validation.
We used the $Ada$ machine at TAMU for the experiments, which is composed of dual socket nodes, each socket containing 10 Intel Xeon E5-2670 v2 (Ivy Bridge-EP) cores (2.5 GHz clock speed) and 64 Gb of memory.
Updated version of Intel compiler and Intel MKL used for the experiments.
\footnote{\url{https://hprc.tamu.edu/wiki/Ada}}

The average relative error is about $60-70\%$ for the basic model, $50\%$ and $45\%$ for the medium model when not-separating and when separating number of divisions from number of flops, respectively, and, $30\%$ for the advanced model.
One reason for the large relative error is that these models make the assumption that most cores behave similarly and that the input matrices are large enough to run the level 3 BLAS routines at full speed.
Although the predictions are not satisfying, our experiments show that there is a good correlation between experimental data and estimated data (of about $0.8$) for most experiments.
Therefore the analytical models can still be used for autotuning purposes to show the trend of the objective function to optimize.

Moreover, we observed that a good set of parameters does not usually correspond to square blocks, although it is usually possible to find a set of parameters involving square blocks whose corresponding runtime is within $70\%$ of the runtime attained by the best parameters.

Furthermore, we observed that there are some patterns in the data and also that the accuracy of the analytical models can vary depending on the region of the search space.
Further investigation is needed however to understand these effects.

\subsection{Multitask learning autotuning}
\label{sss:mla}

In this section, we compare three autotuning approaches: OpenTuner~\cite{opentuner}, HpBandSter~\cite{bohb} and $\MLA$.

In order to be able to compare these methods, given that $\MLA$ requires several tasks while the others only need one, we generated $\NT=50$ tasks following an LHS in the task space.
The performance of a given tuning method on a given task is defined as the lowest computation time of the task yielded by the optimal (best) combination of parameters found by the tuning method.
The stopping criteria of each method on every task is defined as a quota of $20$ evaluations of the objective function (runs of the application) that they are allowed to perform in order to find the optimum.
In the case of $\MLA$, among these $20$ samples, the $12$ first ones are dedicated to the first phase (sampling phase) while the $8$ next ones are dedicated to the second and third phases of the method.
This choice of the number of initial samples to be 3 times the number of parameters is a common practice in the Bayesian optimization literature.
In general, it can affect the runtime and / or accuracy of the $\MLA$ method.
Also, OpenTuner is not able natively to take into account for constraints on the parameters.
Similarly, HpBandSter can handle simple constraints but more advanced constraints cannot be handled, for example, the execution of a routine to check the validity of a combination of parameters.
Thus, we reformulated the parameter space of the application in such a way as all the parameters configurations generated out of this modified space yield valid configurations in the initial parameter space.

\begin{figure}[!h]
    \includegraphics[width=.5\textwidth]{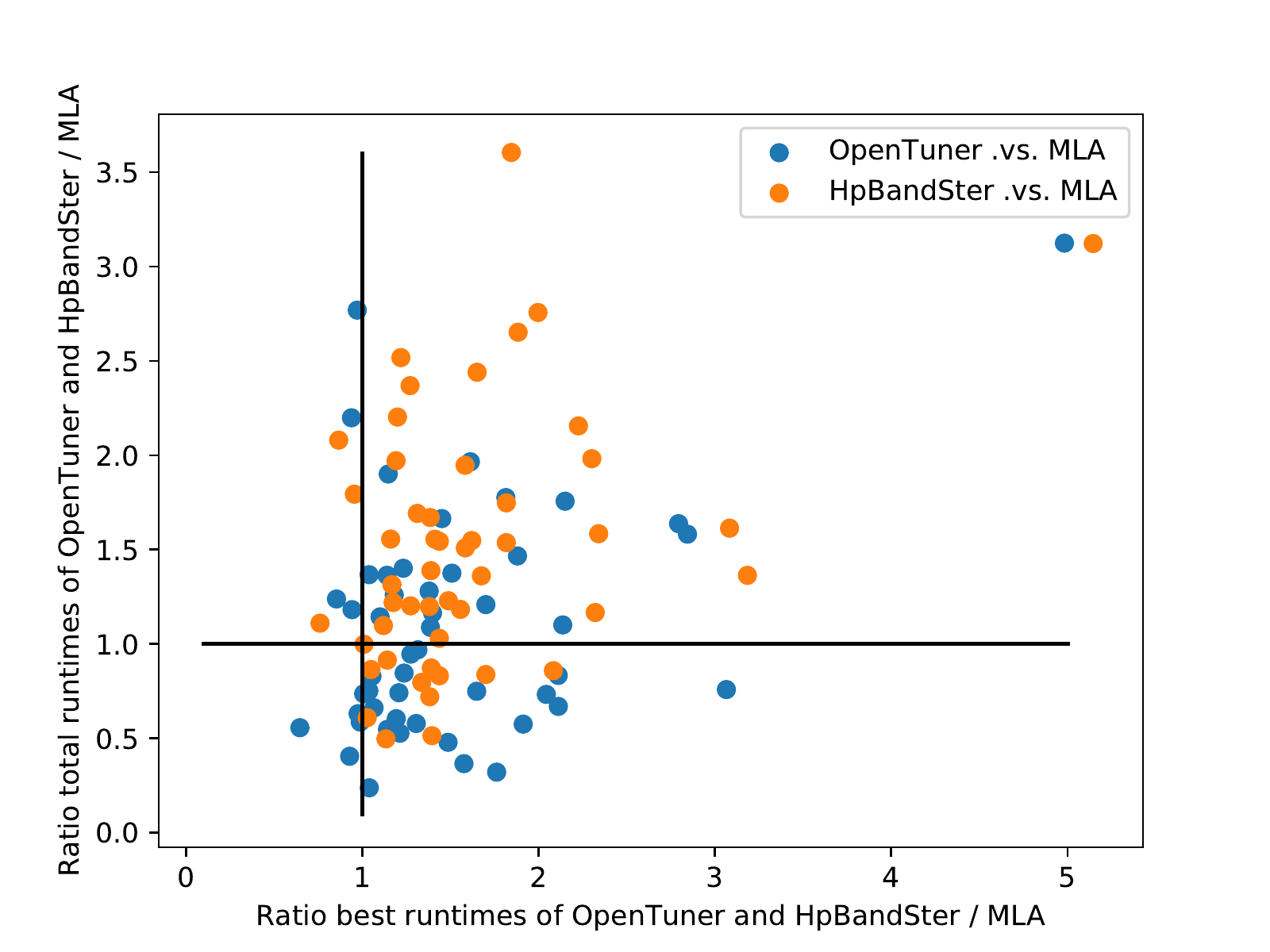}
    \caption{Comparison of $\MLA$ (with Q=20) with OpenTuner and HpBandSter on $50$ different tasks on $128$ nodes of Edison.}
    \label{fig:MLA}
\end{figure}

Figure~\ref{fig:MLA} presents comparative results between $\MLA$, OpenTuner and HpBandSter.
Every point corresponds to a different task.
Moreover, the sizes of the matrices considered is between $1$ and $20000$. 
Furthermore, $128$ compute nodes are used with $24$ cores per node.
Any valid combination of MPI processes and threads is allowed.
The X-axis represents the ratio of best computation time of a task with OpenTuner (blue) and HpBandSter (orange) over that with $\MLA$.
It compares the quality of the tuning of the different methods.
A point at the right of the vertical line means that $\MLA$ finds a better solution than the others, and vice-versa.
The Y-axis represents the ratio of the sum of the runtimes of all the samples generated by OpenTuner and HpBandSter over that with $\MLA$ (excluding the time spent within the search algorithm itself).
It compares the cost (in terms of application evaluations) of the different methods.
A point above the horizontal line means that OpenTuner or HpBandSter are more expensive that $\MLA$, and vice-versa.

Two overall conclusions can be drawn from this figure: 
($i$) $\MLA$ leads to better solutions than OpenTuner and HpBandSter as it leads to better application runtimes in $42$ ($84\%$) and $47$ ($94\%$) cases out of $50$ cases, respectively;
($ii$) $\MLA$ costs less in terms of total application runtime than OpenTuner in $24$ cases and HpBandSter in $38$ cases out of $50$ cases, respectively.

The average $1.5x$ (up to $5x$) improvement of the application runtime using $\MLA$ compared to OpenTuner and HpBandster as observed in the Figure comes at the price of an increased tuning time.
Indeed, although the cost in terms of total runtime of the application on the supercomputer is similar to that of the other tuners, the algorithmic complexity of $\MLA$ is $\O((tn)^3)$, where $t$ is the number of tasks and $n$ the number of samples per task.
In contrast, the other tuners are executed $t$ times with a linear to quadratic algorithmic complexity in $n$.
Indeed, while OpenTuner and HpBandSter require on the order of $1$ to $10$ seconds to make a decision on the next parameter configuration to evaluate, it takes in the order of $155$ seconds for the $\MLA$ model to be built (for the experiment of Figure~\ref{fig:MLA}) and an additional $1$ minute per task to find the best next parameter configuration.
Since OpenTuner and HpBandSter tune every task separately, their total tuning cost is simply their above tuning cost per task times the number of tasks to be tuned.
In contrast, the cost of building the $\MLA$ model increasing cubically with the total number of runs of the application.

Fortunately, several parts of $\MLA$ are amenable to parallelization.
The costly inversion of the covariance matrix in the $\MLA$ model can be achieved in parallel, either using the Cholesky factorization in a multi-thread LAPACK library, or using the distributed-memory Cholesky factorization in the ScaLAPACK library.
Moreover, a second level of parallelism can be taken advantage of for the optimization of the hyper-parameters of the $\MLA$ model, either by using a parallel model-free black-box optimization technique, or by executing the different restarts of a gradient-decent based method in parallel. 
Furthermore, once the $\MLA$ is built, the search for the best next parameter configuration can be done in parallel on the different tasks.
Thus, when tuning exascale applications, the availability of large supercomputers is leveraged, hence alleviating the algorithmic complexity of $\MLA$.

Additionally, algorithmic improvements can be applied to reduce the cost of $\MLA$, potentially at a reduced model quality.
Firstly, the number of latent functions $Q$ is a parameter of the method that inversely impacts the quality and speed of learning of the $\MLA$ model.
Low values of $Q$ lead to faster model learning but decrease the quality of the model.
For instance, in our previous experiment, we have chosen the value of $Q$ to be $20$.
Secondly, sparse approximations exist in the literature that approximate the covariance matrix $K$ (at the heart of the $\MLA$ method) by a low-rank approximation which is much cheaper to deal with.
Again, the lower the rank of the approximation, the worse the quality of the model.
In the GPy package that we rely in the implementation of $\MLA$, the rank is influenced by the choice of a set of {\it inducing points}.
In our previous experiment, we have chosen the number of inducing points as $6\sqrt{tn}$.

Hence, a balance can and must be found in the choice of $Q$ and the rank / inducing points, which would depend on the tuning time that can be afforded together with the relative cost of a run of the application.

\subsection{Transfer learning autotuning}
\label{sss:tla}

Once the $\MLA$ method has been applied on a set of tasks, leading to the construction of a performance model of the application, $\TLA1$ and $\TLA2$ are able to tune new tasks by transferring the knowledge acquired in $\MLA$ to speed up and improve the accuracy of the tuning process.

This section compares the performance of OpenTuner, $\TLA1$ and $\TLA2$.
These three methods leading to better application runtimes than HpBandSter on all the cases tested in this experiment, its results are omitted here.
$10$ new tasks are generated randomly in the task space.
Similarly to the experiment in the previous section, OpenTuner and $\TLA2$ are given a quota of $100$ evaluations of the objective function, $\TLA1$ not needing any.
Also, $50$ points are dedicated to the first phase of $\TLA2$ and $50$ others dedicated to the next phases.
Moreover, the range of size of matrices and hardware configuration is the same as the ones used for in Section~\ref{sss:mla}.

\begin{figure}[!h]
    \includegraphics[width=.5\textwidth]{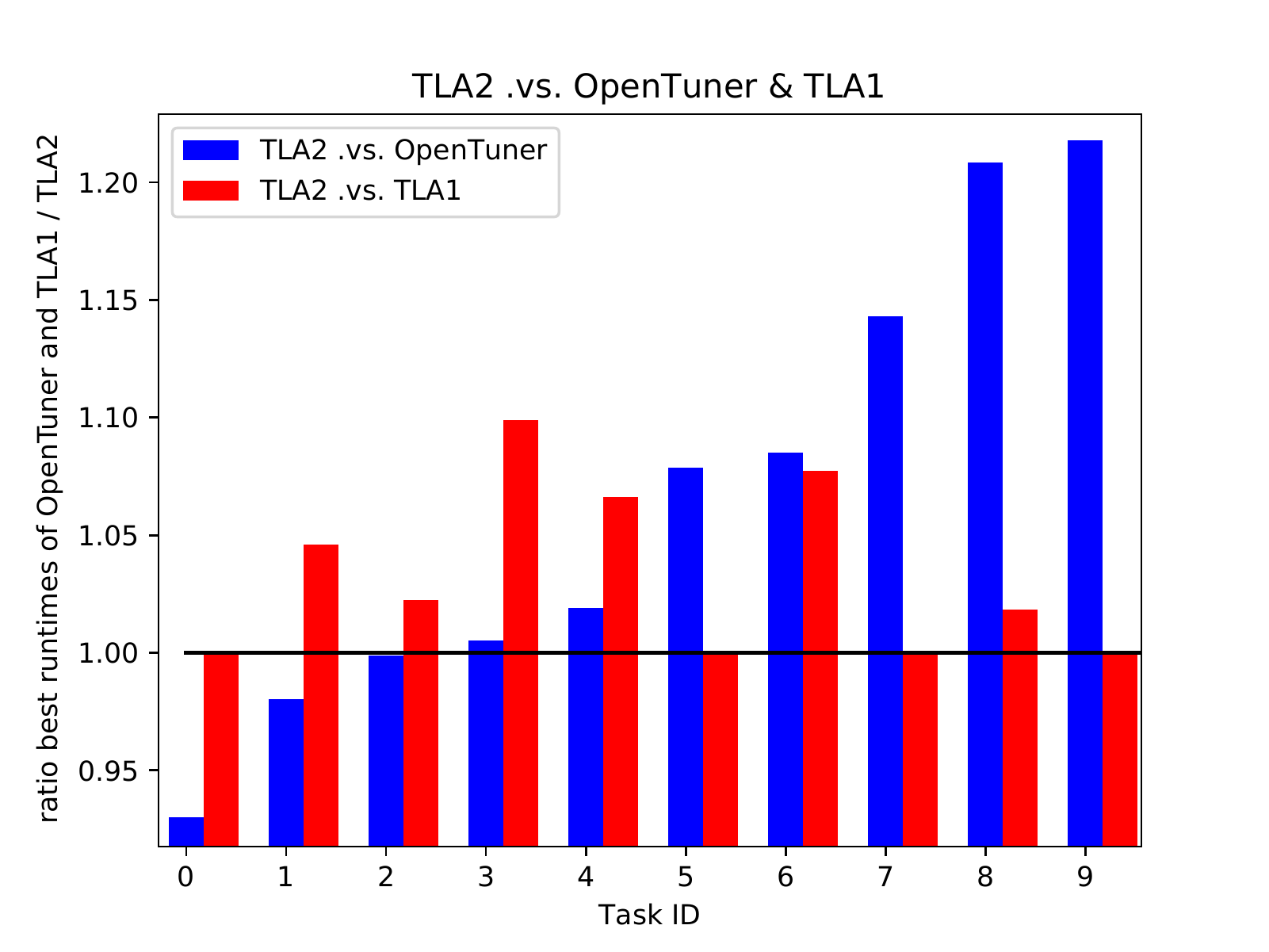}
    \caption{Comparison of $\TLA2$ with OpenTuner and $\TLA1$ on 10 new tasks.}
    \label{fig:tla}
\end{figure}

Figure~\ref{fig:tla} compares the different methods.
The x-axis represents the different tasks.
The y-axis represents the ratio of the best runtime for 
OpenTuner (blue bars) and $\TLA1$ (red bars) over that of $\TLA2$.
The horizontal line at $y=1$ represents the normalized runtime of $\TLA2$.
The tasks are ordered by increasing ratio of best runtime for OpenTuner over $\TLA2$.

In this figure, the performance metric or success metric that we use to compare different tuning methods is the best runtime of the application found by a method on a given problem.
From this figure, we can see that $\TLA1$ is competitive with OpenTuner as they outperform each other in $50\%$ of the cases. 
As $\TLA1$ does not perform any evaluation of the objective function, but guesses the optimal parameter configurations through a model prediction, this shows that the transfer of the learning from the model of $\MLA$ is successful.
Moreover, $\TLA2$ outperforms $\TLA1$ in $6$ cases, and performs similarly in the other cases.
This result is not surprising as $\TLA2$ builds on top of $\TLA1$, and incorporates the result of $\TLA1$ in its samplings.
Finally, $\TLA2$ outperforms OpenTuner in $7$ cases out of $10$ and is
outperformed in $2$ cases out of $10$.
$\TLA2$ is thus a viable competitor to state-of-the-art autotuning methods.

\subsection{Semi-Exhaustive search results}
\label{sss:exost}

Figure~\ref{fig:semiexhost} shows the performance of the $\PDGEQRF$ routine of ScaLAPACK on the {\it Edison} computer, at {\it NERSC}, for two problem configurations, that is, a different $\m$, $\n$, $\nodes$. The computer being the same, $cores=24$.
Figure~\ref{fig:2e3} corresponds to $\m=\n=2000$ and $\nodes=1$ (shared-memory setting), while Figure~\ref{fig:1e4} corresponds to $\m=\n=10000$ and $\nodes=128$ (distributed-memory setting).
On both its sub-figures, the X and Y axes correspond to $\mb$ and $\nb$, respectively. The Z axis represents the computation time.  Given that the remaining parameters cannot be represented on a 3D graphic, every layer in the lasagna-like graphic correspond to a different combination of all of the other parameters ($\nproc$, $\nth$, $\p$ and $\q$).
Also, in every layer, the lighter the color, the higher the value of the objective function, while the darkest colors correspond to the regions of interest where the optima lies.
The global optima correspond to the darkest region(s) at the lowest layer(s).
The best parameters for each problem are described in Table~\ref{tab:bstparams}:

\begin{figure}[!h]
    \begin{subfigure}[!h]{0,5\textwidth}
        \includegraphics[width=\textwidth]{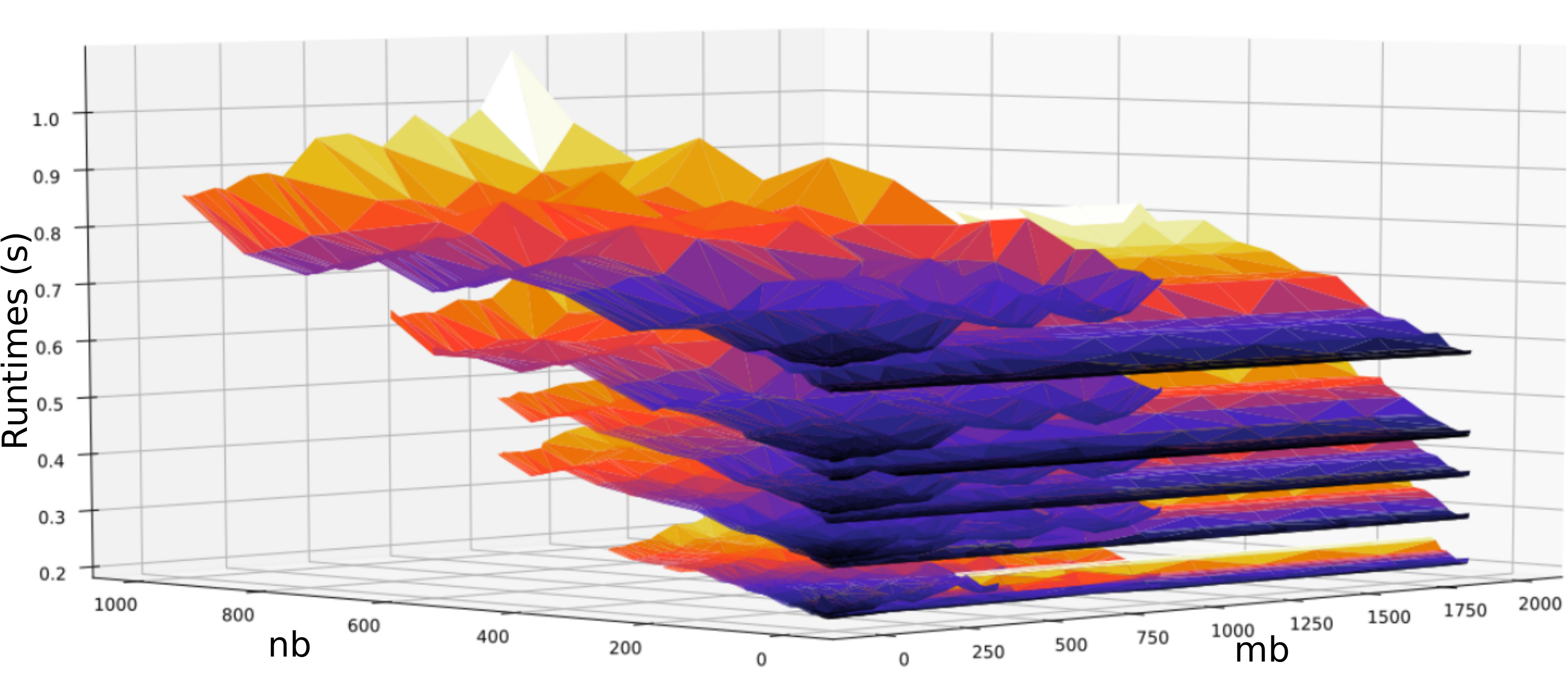}
        \caption{2000x2000 matrix on 1 node}
        \label{fig:2e3}
    \end{subfigure}
    \begin{subfigure}[!h]{0,5\textwidth}
        \includegraphics[width=\textwidth]{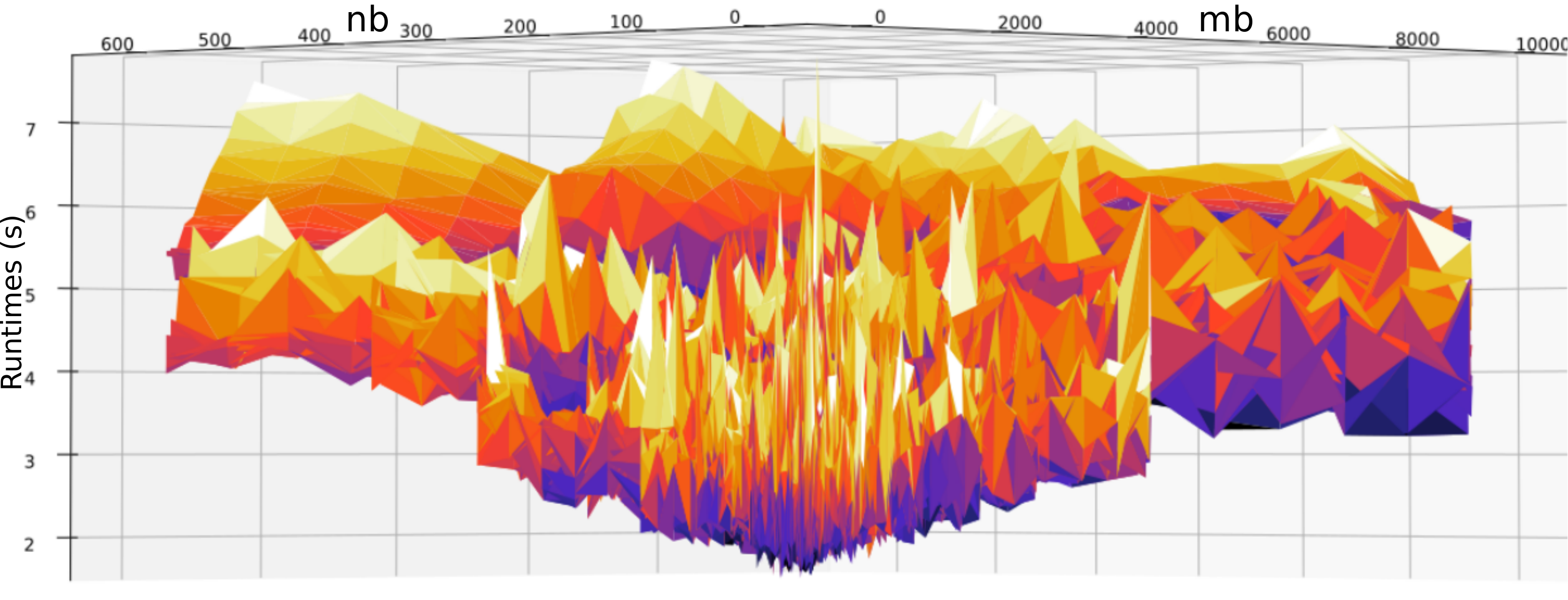}
        \caption{10000x10000 matrix on 128 nodes}
        \label{fig:1e4}
    \end{subfigure}
    \caption{Grid search on two different settings of a ScaLAPACK QR factorization.}
    \label{fig:semiexhost}
\end{figure}

\begin{table}[!h]
    \scalebox{1.0}{
    \begin{tabular}{|@{\hskip3pt}r@{\hskip3pt}r@{\hskip3pt}r@{\hskip3pt}r@{\hskip3pt}|@{\hskip3pt}r@{\hskip3pt}r@{\hskip3pt}r@{\hskip3pt}r@{\hskip3pt}r@{\hskip3pt}r@{\hskip3pt}|@{\hskip3pt}r@{\hskip3pt}|}
        \hline
        \multicolumn{4}{|c|}{Problem} & \multicolumn{6}{|c|}{Best parameters} & Time (s) \\
        \hline
        m & n & nodes & cores & mb & nb & nth & nproc & p & q & \\
        \hline
        2000 & 2000 & 1 & 24 & 4 & 8 & 1 & 24 & 2 & 12 & 0.20 \\ 
        10000 & 10000 & 128 & 24 & 1 & 26 & 1 & 3072 & 16 & 192 & 1.61 \\ 
        \hline
    \end{tabular}
    }
    \caption{Best parameters yielding the best runtime found by grid search on two different settings of a ScaLAPACK QR factorization.}
    \label{tab:bstparams}
\end{table}

From these results, several conclusions can be drawn.
Firstly, we can see that regions of interest exist, where the optima should be, and can be discovered by the optimization methods.
Roughly speaking, they are located, in the first case, at small values of $\nb$, while in the second case, at small values of both $\mb$ and $\nb$.
Secondly, the different lasagna layers are well separated in the first case while somehow interleaved in the second case.
This last scenario makes it harder for an autotuner to figure out the optimal parameter configuration.
Finally, in the second scenario, the regions where the optimum seems to be have a great amount of variability or noise in them, while non-interesting regions undergo
much less variability.
This can be explained by the fact that, small values of $\mb$ and $\nb$ lead 
to several small messages to be transferred over the network while other values of 
these parameters lead to fewer but larger messages to be transmitted.
The first case is much more prone to be affected by network noise than the second case.
Indeed, when large scale applications running on supercomputers, the applications
share the computer network with each other.
This phenomenon is a serious challenge for autotuning at the exascale.


If we consider that the best runtime found by means of grid search is close to optimal, it is possible to compare the other methods in terms of absolute performance (instead of relative).
Table~\ref{tab:compmeths} offers such a comparison between grid search, OpenTuner, \TLA1 and \TLA2 on the tuning of a single task (matrix of size 500-by-500).
{\it Budget} represents the number of runs of the application allowed (or necessary) for a specific method.
{\it mb, nb, nth, nproc, p and q} represent the values of the parameters for the optimal configuration found by a given method, while the column {\it Best Time} shows the corresponding best runtime.
OpenTuner finds the least good solution (compared to the other methods) when given a budget of $100$ and a good solution when given a budget of $1000$ (impractical in an exascale setting).
An approach such as that of OpenTuner is thus not adequate for tuning exascale applications, where the budget is limited.
However, \TLA1 (requiring 0 runs) and \TLA2 (with a budget of $100$) find competitive solutions, even with such a low budget, making them more suitable candidates for tuning exascale applications.

\begin{table}[!h]
    \scalebox{1.0}{
    \begin{tabular}{|@{\hskip3pt}l@{\hskip3pt}|@{\hskip3pt}r@{\hskip3pt}|@{\hskip3pt}r@{\hskip3pt}|||r@{\hskip3pt}r@{\hskip3pt}r@{\hskip3pt}r@{\hskip3pt}r@{\hskip3pt}@{\hskip3pt}r@{\hskip3pt}|}
        \hline
        Method & Budget & Time (s) & \multicolumn{6}{|c|}{Best parameters} \\
        \hline
                    &      &         &  mb & nb & nth & nproc & p &  q  \\
        \hline                      
        OpenTuner   &  100 & 7.00e-3 & 470 & 30 &   3 &     8 & 1 &  8 \\
        OpenTuner   & 1000 & 5.30e-3 &  88 & 12 &   1 &    24 & 1 & 24 \\
        \TLA1       &    0 & 5.73e-3 & 244 & 11 &   1 &    24 & 1 & 24 \\
        \TLA2       &  100 & 5.67e-3 & 296 & 11 &   1 &    24 & 1 & 24 \\
        Grid search & 8192 & 5.15e-3 & 500 &  8 &   1 &    24 & 1 & 24 \\
        \hline
    \end{tabular}
    }
    \caption{Best parameters and best runtime found by grid search, OpenTuner, \TLA1 and \TLA2 for a ScaLAPACK QR factorization of a matrix of size 500-by-500 on 1 node of Edison with 24 cores.}
    \label{tab:compmeths}
\end{table}

%
%
%
%
%
%

\section{Conclusion}
\label{sec:cncl}


This work presents the use of multitask and transfer learning for the purpose of autotuning.
The core ideas are:
$(i$) multitask learning: instead of tuning an application on different input problems independently one from the other, the knowledge of all tasks can help tune every task faster,
$(ii)$ transfer learning: the knowledge accumulated from tuning previous tasks can be exploited to speed up the tuning of the new tasks.

The methods in this paper are intended to be used as follows.
The user of an application would tune this application on an initial set of representative problems.
This expensive tuning phase would be done once, through $\MLA$, in order to create a model of the objective function to optimize.
Then, every time the application is to be tuned on a new task, either $\TLA1$ or $\TLA2$ would be used at a much reduced cost, the choice depending on the accuracy of the solution required or the amount of tuning time and computational resources available.
If too many new tasks have been tuned through $\TLA2$, the user might want to re-build a new performance model using $\MLA$ from scratch, as a large amount of data is available that was not during the initial build of the model.

The experimental results show that $\MLA$ outperforms OpenTuner on $66\%$ of the cases they are compared on, with an application runtime improvement of up to $40\%$.
Moreover, the run-free $\TLA1$ and the low-cost $\TLA2$ methods find configurations whose runtimes outperform or are very competitive with those found by OpenTuner and HpBandSter.
Furthermore, when the budget for autotuning (number of allowed runs of the application) is low, for instance, due to a high cost of the target application (e.g. exascale applications), our methods are more suitable than the other state-of-the-art methods,
which are not necessarily designed for this kind of use case.


This work presents a new view to autotuning that opens the doors to new challenges and research.
Future directions include the use of:
$(i)$ more complex models (e.g. {\it deep Gaussian processes}) to better take into account the non-stationarity, discontinuity and heavy noise 
which are characteristic behaviours of the objective functions measured on exascale applications (See Figure~\ref{fig:1e4}),
$(ii)$ different models (e.g. Gaussian processes with random forests) to better handle the case of categorical parameters, and
$(iii)$ model approximations (e.g. {\it sparse Gaussian processes}) in order to scale the optimization algorithms themselves.

\ignore{
\section*{Acknowledgment}

This research was supported by the Exascale Computing Project (17-SC-20-SC), a collaborative effort of the U.S. Department of Energy Office of Science and the National Nuclear Security Administration.
We used resources of the National Energy Research Scientific Computing
Center (NERSC), a U.S. Department of Energy Office of Science User Facility
operated under Contract No. DE-AC02-05CH11231.
}

\bibliographystyle{abbrv}
\bibliography{bib}


\end{document}